\documentclass[11pt,a4paper]{article}
\PassOptionsToPackage{numbers, compress}{natbib}

\usepackage[preprint]{nips2018}

\usepackage[utf8]{inputenc} %
\usepackage[T1]{fontenc}    %
\usepackage{hyperref}       %
\usepackage{url}            %
\usepackage{booktabs}       %
\usepackage{amsfonts}       %
\usepackage{nicefrac}       %
\usepackage{microtype}      %
\usepackage{graphicx}
\usepackage{times}
\usepackage{latexsym}
\usepackage{amssymb}
\usepackage{url}
\usepackage{graphicx}
\usepackage{todonotes}
\usepackage{float}
\usepackage{booktabs,multirow}
\usepackage[font={footnotesize}]{caption}
\usepackage{bibunits}

\defaultbibliography{emnlp2018}
\defaultbibliographystyle{acl_natbib_nourl}
  \newcommand*\samethanks[1][\value{footnote}]{\footnotemark[#1]}

\title{Adversarial Gain}

\author{Peter Henderson$^1$, Koustuv Sinha$^1$\thanks{\hspace{1em}Authors contributed equally.}, Rosemary Nan Ke$^2$\samethanks, Joelle Pineau$^1$\\$^1$ Mila, McGill University\\$^2$ Mila, Polytechnique Montr\'eal}

\begin{document}
\maketitle
\begin{abstract}
Adversarial examples can be defined as inputs to a model which induce a mistake -- where the model output is different than that of an oracle, perhaps in surprising or malicious ways. Original models of adversarial attacks are primarily studied in the context of classification and computer vision tasks. 
While several attacks have been proposed in natural language processing (NLP) settings, they often vary in defining the parameters of an attack and what a successful attack would look like.
The goal of this work is to propose a unifying model of adversarial examples suitable for NLP tasks in both generative and classification settings.
We define the notion of adversarial gain: based in control theory, it is a measure of the change in the output of a system relative to the perturbation of the input (caused by the so-called adversary) presented to the learner.
This definition, as we show, can be used under different feature spaces and distance conditions to determine attack or defense effectiveness across different intuitive manifolds.
This notion of adversarial gain not only provides a useful way for evaluating adversaries and defenses, but can act as a building block for future work in robustness under adversaries due to its rooted nature in stability and manifold theory.
\end{abstract}

\begin{bibunit}
\section{Introduction}

The notion of \emph{adversarial examples} has seen frequent study in recent years~\cite{szegedy2013intriguing,goodfellow2014explaining, miyato2016adversarial,jia2017adversarial,  elsayed2018adversarial}. The definition for adversarial examples has evolved from work to work\footnote{See Supplementary Material for definitions in prior work}. However, a common overarching definition\footnote{\href{https://blog.openai.com/adversarial-example-research/}{https://blog.openai.com/adversarial-example-research/}} characterizes adversarial examples as \textit{``inputs to machine learning models that an attacker has intentionally designed to cause the model to make a mistake.''}

In such a context a mistake can be defined such that a model's output $f(x)$ differs from the output of a set of oracle (or optimal) models $f^*(x)$. In some cases the oracle output is known and this definition is sufficient. One such example is in the case of malware detection~\cite{grosse2016adversarial}. A target sample is known to be malware, but can be disguised -- without the possibility of changing its ground truth role as malware -- to cause a malware detection model to classify it as a safe sample.

However, in some cases, the optimal output given the perturbation or generated sample is unavailable or ambiguous. Furthermore, evaluation methods of the output may not be descriptive enough as an alternative for assessing performance under an adversary -- as in dialogue~\cite{liu2016not} or translation~\cite{callison2006re}. 

To circumvent the lack of availability of an oracle model or descriptive evaluation metric, various works have made distance-based assumptions surrounding adversarial examples. A known sample is perturbed by a constrained amount such that within the constraint the output of the model output should be unchanged.

In NLP tasks, various works attempt to preserve meaning (and thus ensure that the oracle output should be unchanged) by constraining operations, such as only replacing words with synonyms \cite{alzantot2018generating,papernot2016crafting,cheng2018seq2sick,zhao2017generating,ebrahimi2017hotflip}. However, such constraints are task-dependent, often difficult to specify, and not necessarily guaranteed. There can be cases where a model output may correctly change its output within some constrained radius perturbation (e.g., if a sentence is on the border between two sentiments, a small change may cause the classifier to make a valid shift). In fact, in a survey conducted by~\citet{jia2017adversarial} about their generated adversarial examples, it was found that humans -- a proxy for the oracle in this setting -- sometimes did change their answer under the perturbed noise.

Finally, in text generation settings the notion of what constitutes a mistake varies from work to work. \citet{miyato2016adversarial, papernot2016crafting,cheng2018seq2sick,zhao2017generating} measure an adversary's effectiveness in generating a target word or sequence; \citet{zhao2017generating} create an adversary which successfully causes a model to omit words; \citet{cheng2018seq2sick} introduce a measure of success where the model outputs text that has no overlap with its original output; \citet{ebrahimi2017hotflip} measure success rate as a function of the decrease in BLEU score beyond some threshold.

\section{Adversarial Gain}

To account for the lack of guarantees in perturbation constraints, the sometimes ambiguous notion of a ``mistake'' by a model, and the unknown oracle output for a perturbed sample, we propose the unified notion of \emph{adversarial gain}.
We draw from incremental $L_2$-gain in control theory~\cite{van2000l2} as inspiration and define the adversarial gain as:
\begin{equation}
\hat{\beta}_{adv} \le \frac{ D_{out}(\phi_{out}(f(x)), \phi_{out}(f(x_{adv}))) }{D_{in}(\phi_{in}(x), \phi_{in}(x_{adv}))},
\end{equation}
such that $x$ is a real sample from a dataset, $x_{adv}$ is an adversarial example according to some attack targeting the input $x$, $x \ne x_{adv} \forall (x, x_{adv}) \in X$, $f(x)$ is the learner's output, $\phi_{in}, \phi_{out}$ is a feature transformation for the input and output respectively, and $D_{in}, D_{out}$ are some distance metrics for the input and output space respectively. $\beta_{adv}$ indicates per sample adversarial gain and $\hat{\beta_{adv}}$ is an upper bound for all samples $X$. 

We do not assume that a model's output should be unchanged within a certain factor of noise as in \citet{raghunathan2018certified,bastani2016measuring},
 rather we assume that the change in output should be proportionally small to the change in input according to some distance metric and feature space. Similar to an $L_2$ incrementally stable system, the goal of a stable system in terms of adversarial gain is to limit the perturbation of the model response according to a worst case adversarial input $x_{adv}$ relative to the magnitude of the change in the initial conditions. Since various problems place emphasis on stability in terms of different distance metrics and feature spaces, we leave this definition to be broad and discuss various notions of distance and feature spaces subsequently.

 This notion holds for both cases where an oracle is known and unknown, for both generative and discriminative settings, and for continuous and discrete spaces. Furthermore, this allows for an adversary to make arbitrarily large changes in the input space, so long as the change causes proportionally large an instability in the output space. In cases where the oracle output is known (e.g., we know that a malware should be classified as such), a traditional metric, such as model accuracy across adversarial examples, can be used in conjunction with adversarial gain. In these settings, gain can provide additional information about the vulnerable space of inputs, similarly to the manifold space as used in~\citet{wu2017manifold}. Additional properties are discussed in Supplementary Material.

\begin{table*}[!htbp]
    \centering
    {\small{\begin{tabular}{p{12.6cm}}
         \hline
         \textbf{Input:} leading season scorers in the bundesliga after saturday 's third-round games \textbf{(periods)} : UNK\\
         \textbf{Original output:} games standings | \textbf{Adversarial output:} Scorers after third-round period\\
         $\beta_{adv}=9.5$, $D_{in}=0.05$, $D_{out}=0.5$, Word-overlap: 0\\
         \hline
         \textbf{Input:} palestinian prime minister ismail haniya insisted friday that his hamas-led \textbf{(gaza-israel)} government was continuing efforts to secure the release of an israeli soldier captured by militants .\\\
         \textbf{Original output:} hamas pm insists on release of soldier | \textbf{Adversarial output:} haniya insists gaza truce efforts continue\\
        $\beta_{adv}=4693.82$, $D_{in}=0.00$, $D_{out}=0.46$, Word-overlap: 1\\
        \hline
        \textbf{Input:} south korea \textbf{(beef)} will \textbf{(beef)} play for \textbf{(beef)} its \textbf{(beef)} third straight olympic women 's \textbf{(beef)} handball gold medal when \textbf{(beef)} it meets denmark saturday \textbf{(beef)}\\
        \textbf{Original output:} south korea to meet denmark in women 's handball | \textbf{Adversarial output:} beef beef beef beef beef beef beef up beef\\
        $\beta_{adv}=3.59$, $D_{in}=0.15$, $D_{out}=0.55$, Word-overlap: 0\\
         \hline
    \end{tabular}}}
        \vspace{1mm}
    \caption{Adversarial examples for text summarization using \cite{cheng2018seq2sick}. The bold words are those which modify the original sentence. Brackets indicate an addition, parenthesis indicate replacement of the preceding word. An $\epsilon=1^{-4}$ is added to the denominator to avoid division by 0 in this case. $D_{in}, D_{out}$ both in terms of InferSent distance.}
    \label{tab:qualit2}
\end{table*}

\subsection{Bootstrapping the Real Data Gain}

Since adversarial gain on its own doesn't necessarily indicate a mistake, we must also determine what is an unusual amount of gain. That is, at what point has the model begun to generate likely incorrect outputs. To do this, we can bootstrap some rough bounds from the known data. That is for any two batches ($M_1, M_2$) of data randomly sampled from the known data such that $M_1 \cap M_2 = \varnothing$, we generate a set of bootstrap samples:
\begin{equation}
\beta_{M,real} = \frac{ D_{out}(\phi_{out}(f(x_1)), \phi_{out}(f(x_{2})))}{D_{in}(\phi_{in}(x_1), \phi_{in}(x_{2}))}, 
\end{equation}
where $x_1,y_2 \in M_1, x_2,y_2 \in M_2$, and $\hat{\beta}_{M,real}$ indicates an upper bound.

From these gain samples, we can estimate some bounds on the average point-wise gain of the real data using the bootstrap~\cite{efron1986bootstrap}. We refer to this bootstrap estimate as $\beta_{real}$, or the ``real'' gain. If an adversarial example has a gain exceeding the bootstrap estimate, it is more likely that the model in fact made a mistake due to an adversary. That is, given some level of change in input, has the output shifted into a significantly different space than what is typical in known data.

\subsection{Distance Metrics and Feature Spaces}

Our definition of adversarial gain depends crucially on the definition of distance metrics for both the input and output spaces.

\subsubsection{Distance Metrics in NLP}

There are many distance metrics relevant for NLP tasks as discussed by~\citet{van2012domain}. These include divergences in probability distributions (e.g., Jensen-Shannon divergence), semantic similarity~\cite{mihalcea2006corpus}, count-based metrics (word overlap, BLEU score, etc.), and various string kernels~\cite{lodhi02}.

For NLP input spaces, while count-based metrics provide some signal, they are often lacking as evaluation and distance measures as discussed in~\cite{callison2006re} and seen in Section~\ref{sec:exp}. Using semantic similarity or cosine similarity has been used in~\citet{henderson2017ethical} for investigating adversarial examples in dialogue. It comes with the intuition that similar linguistic samples should be closer together. However, measuring semantic similarity can be difficult due to the language understanding required and often needs a well-defined feature space. 

On the output side, for classification tasks, such as sentiment classification~\cite{socher2013recursive}, it is possible either to use a step-wise function (1 if classification changes, 0 otherwise) or a divergence. As \citet{wu2017manifold} do, the latter suggests that ``confident regions of a good model should be well separated'', but in the context of adversarial gain should be proportional to the input distance for reduced gain. Moreover, proper use of uncertainty or distribution modeling can be shown to protect against adversarial attacks~\cite{bradshaw2017adversarial}, and thus evaluating the gain in terms of probabilistic divergences may be desirable.

\begin{figure}
    \centering \includegraphics[width=.6\textwidth]{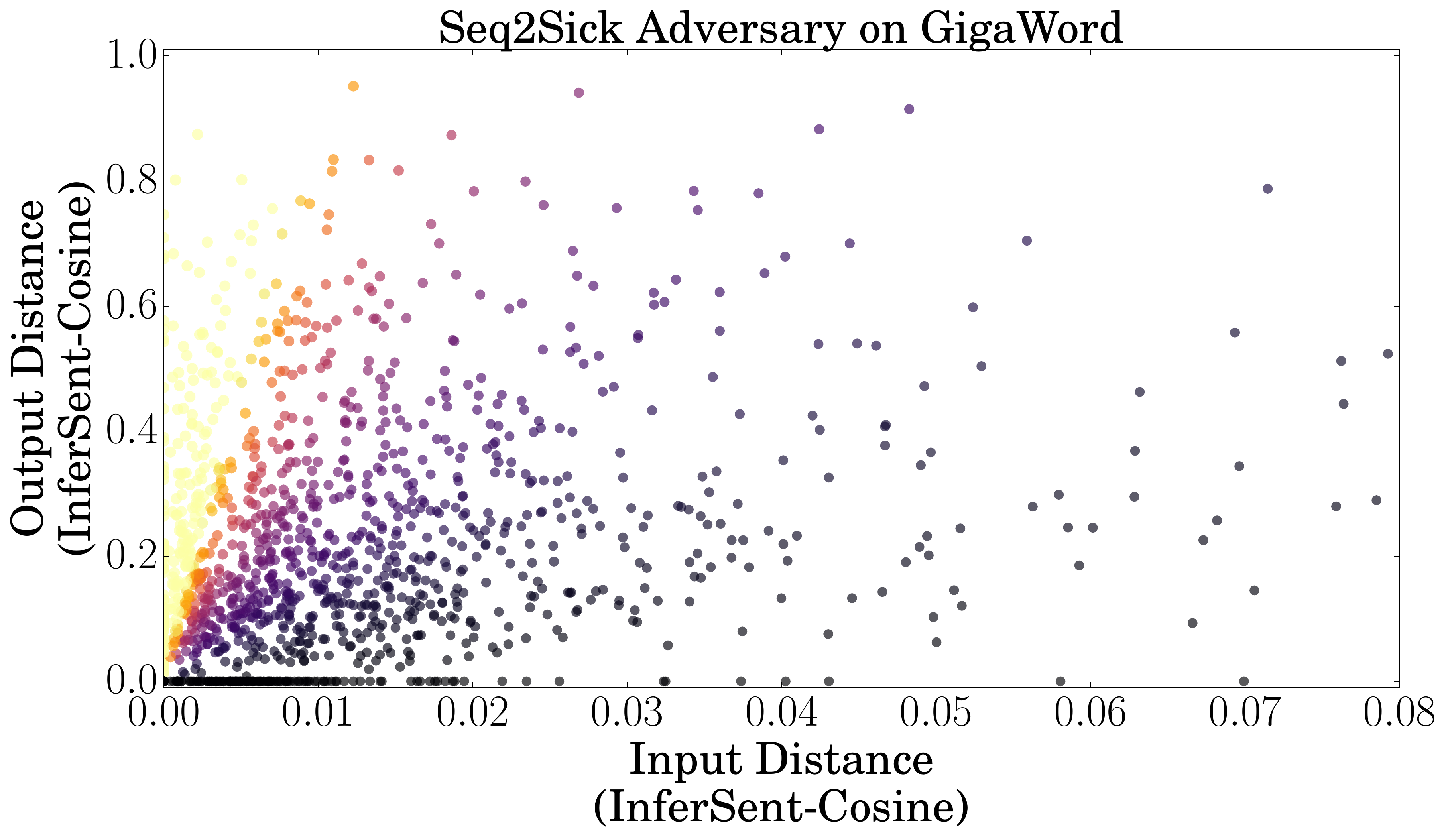}
    \caption{The distribution of adversarial examples in text summarization tasks. Warmer colors (reg, orange, yellow, respectively) indicate higher gain values.}
    \label{fig:gaingraph}
\end{figure}

\subsubsection{Feature Spaces}
To measure semantic similarity of text, various encoding methods have been developed which transform the text into a vector space~\cite{kiros2015skip,conneau2017supervised,yang2018learning,cer2018universal}. Using the cosine similarity in conjunction with such an embedding space can ensure that similar text will be closer together. Here, we use the InferSent embeddings~\cite{conneau2017supervised} as the primary form of measuring semantic similarity. Adversarial gain can be measured across different feature spaces (and thus different manifolds). However, another appropriate method may be to learn a specific embeddings (feature) space for the problem at hand similarly to~\citet{yang2018learning} since well-generalized embedding spaces are difficult to create~\cite{dasgupta2018evaluating}. By learning a feature space which ensures a well-defined distance-based correlation between inputs and outputs, the distance assumption can more accurately measure whether an adversarial attack falls in the gain range where a mistake is more likely. This follows manifold-based work as in~\citet{wu2017manifold,lamb2018fortified}.

\subsection{A Note on Human Perception}
A common debate regarding adversarial examples is whether they should be perceivable by humans. Many works cite perception in their definition of adversarial examples or run surveys determining whether humans were able to perceive the change~\cite{shaham2015understanding,hein2017formal,ilyas2017robust,jia2017adversarial}. However, \citet{elsayed2018adversarial} contest that the use of perception in the definition is incorrect because then humans would not be susceptible to adversarial examples -- and they claim later on that humans in fact are susceptible under some constrained conditions. In the setting of adversarial gain, human perception is not a strict condition. However, human perception plays a relation to the oracle. In many tasks, human perception is used as a proxy for an oracle model. For example, in image classification, datasets are generated from what humans perceive to be the label rather than some verified ground truth. In the context of adversarial gain, it is possible that humans are susceptible to certain high gain samples such as the perceived colour of ``the dress''~\cite{winkler2015asymmetries}. This satisfies the properties set forth by~\citet{elsayed2018adversarial}. However, it also allows for the accounting of human perception. By measuring the adversarial gain bounds of humans across distance metrics, it may be possible to build a better picture of expected model performance in many ambiguous settings where we use humans as proxies for an oracle model (e.g., dialogue, text summarization, sentiment analysis).

\subsection{A Note on Generative Adversarial Examples}

In the case where samples are not perturbed, but rather generated from scratch as in~\cite{zhao2017generating,xiao2018generating,2018arXiv180507894S}, there is no original sample to be compared against. In this case, we can think about the use of our latent feature space $\phi$ and find the nearest known neighbourhood of examples within that feature space. These can be used as a reference point for evaluating the gain of the adversarial example. 
This can be applied to perturbed adversarial gain as well, but is computationally much more intensive. Finding high gain samples in such a way may allow for the discovery of unknown regions of space where more real samples are needed or decision boundaries and certainty gradients must be adjusted.

\subsection{A Note on Targeted Attacks}

We do not explicitly consider targeted attacks in our main definition of adversarial gain. However, because of the distance based formulation, it is simple to do so.
A targeted attack can be thought of in two ways: (1) inducing a model to generate a certain output (even if it's not wrong); (2) inducing a model to make a mistake in a particular way which generates a certain output. We posit that some prior literature actually examines the first case. \citet{cheng2018seq2sick}, for example, use an indicator function which determines if a certain set of words exists in an output sequence. One example of a success for inducing the words ``Hund sizst'' in a machine translation task that is provided in~\cite{cheng2018seq2sick} is:

\small{
\begin{quote}
  \textbf{SOURCE INPUT SEQ:} A TODDLER IS COOKING WITH ANOTHER PERSON.
  
\textbf{ADV INPUT SEQ:} A dog IS sit WITH ANOTHER UNK.

\textbf{SOURCE OUTPUT SEQ:} EIN KLEINES KIND KOCHT MIT EINER ANDEREN PERSON.

\textbf{ADV OUTPUT SEQ:} EIN Hund sitzt MIT EINEM ANDEREN UNK.
\end{quote}}

It is clear in this case that the model does not necessarily make a mistake, but rather changes in the input to induce a certain output that is a correct translation. While this is an interesting problem and approach, the model is still performing as expected in this case. We instead, can formulate targeted adversarial gain in the context of the latter where we need to have a notion of distance to a known sample to approximate incorrect behaviour. We can define gain as the difference between two distances, that of the original sample to the target sample and the adversarial sample to the target sample. This forms a sort of cost-to-go function. That is for a target, a large adversarial gain corresponds to the closest input change to reach a certain target output space. In terms of classification tasks, this may have interesting properties related to decision boundaries, but we consider it out of scope for our experiments. 

\section{Experiments}

We aim to study empirically whether adversarial gain is suitable as a unified notion in both generative and discriminative NLP settings.  We run  experiments on text summarization and sentiment classification based on existing open-source constrained adversarial attacks, and evaluate whether adversarial gain offers a relevant characterization.

\begin{table}[!htbp]
    \centering
    \begin{tabular}{|c|c|c|}
    \hline
             Metrics & $\beta_{real}$ & $\beta_{adv}$\\
             \hline
    \multicolumn{3}{|c|}{Sentiment Classification}\\
    \hline
         IS + JS& 0.85 (0.79, 0.91) & 13.75 (-1.93, 25.32)\\
         IS + Step & 1.18 (1.10, 1.27) & 22.6 (-3.96, 42.5)\\
         WD + Step & 0.018 (0.016, 0.019) & 0.241 (0.227, 0.255)\\
         WD + JS & 0.008 (0.007, 0.008) & 0.121 (0.115, 0.127)\\
         \hline
             \multicolumn{3}{|c|}{Text Summarization}\\
             \hline
         IS + IS & 2.174 (2.14, 2.20) & 134.62 (102.31, 163.05)\\
         \hline
    \end{tabular}
    \vspace{1mm}
    \caption{We provide the bootstrap average with confidence bounds across 10k bootstrap samples. To avoid division by $0$, we add an $\epsilon=1^{-4}$ to the denominator of the gain. WD indicates the number of words that word added or changed. IS indicates the InferSent cosine distance. Step indicates 1 if the class label changed, 0 otherwise. }
    \label{tab:metrics}
\end{table}

\subsection{Generative Tasks: Text Summarization}
\label{sec:exp}

For text summarization we use the GigaWord dataset~\cite{rush2015neural,graff2003english,napoles2012annotated}, subset of holdout test data, pretrained model, word embeddings, and attack vector as used by \citet{cheng2018seq2sick}. We use InferSent embeddings, and cosine distance to measure the distance on both inputs and outputs.

The resulting bootstrap estimate average gain can be seen in Table~\ref{tab:metrics} and the distribution of change caused by the adversarial attack can be visualized in Figure~\ref{fig:gaingraph}. It is clear that the attack does induce changes in meaning on average according to the InferSent embeddings, but there are also low-gain samples where the attack must make large changes in the input space to cause a significant change in output. \citet{cheng2018seq2sick} measure the success of an attack if there is no word overlap in the changed output. While this does provide some information, it may be the case that the model is still technically correct in its performance even with no overlap. The first example in  Table~\ref{tab:qualit2} demonstrates such a scenario. Adversarial gain in a feature space such as InferSent, however, provides a more refined notion of change. Furthermore, the second sample in Table~\ref{tab:qualit2} demonstrates a high gain due to change in meaning even though there is word overlap. Lastly, in a case where there is no overlap in the outputs due to a large number of changes to the input meaning, the notion of adversarial gain gives the model some leeway (if the input is drastically changed it's likely okay to change the output). As seen in Table~\ref{tab:metrics}, on average these scenarios fall outside of the typical bound of the real data indicating some level of attack effectiveness, thus showing that adversarial gain provides a decent notion of the effectiveness of an attack and susceptibility of the model to attack.

\subsection{Discriminative Tasks: Sentiment Classification}
\begin{table}[!htbp]
    \centering
    {\small{\begin{tabular}{p{7.2cm}}
         \hline
         a benign but forgettable sci fi diversion \textbf{[fiorentino brio]}\\
         $f(x)= ( 0.98, 0.02)$, $f(x_{adv})=(0.02, 0.98)$\\
         $\beta_{adv}=\infty$, $D_{in}=0.0$, $D_{out}=0.60$\\
         \hline
         the transporter is as lively and as fun as it is unapologetically dumb \textbf{(ineffective)}\\
         $f(x)= ( 0.01, 0.99)$, $f(x_{adv})=(0.99, 0.01)$\\
         $\beta_{adv}=4.94$, $D_{in}=0.13$, $D_{out}=0.64$\\
         \hline
         ranks among willams ' best screen work \textbf{[cram cheesy]}\\
         $f(x)= ( 0.00, 1.00)$, $f(x_{adv})=(0.66, 0.34)$\\
         $\beta_{adv}=1.34$, $D_{in}=0.22$, $D_{out}=0.31$\\

         \hline
    \end{tabular}}}
    \vspace{1mm}
    \caption{Adversarial examples for sentiment classification using~\citet{ebrahimi2017hotflip}. The bold words are those which modify the original sentence. Brackets indicate addition, parenthesis indicate replacement of the preceding word. $D_{in}$ is the InferSent distance. $D_{out}$ is the JS divergence.}
    \label{tab:qualit}
\end{table}

Next, we examine a sentiment classification task using the SST2 dataset \cite{socher2013recursive}, pre-trained convolutional neural network model, and single word flip attack as provided by \citet{ebrahimi2017hotflip}. We use a step-wise function and the JS divergence as distance metrics on the output. We use InferSent embeddings and word distance (number of different words) as measures on the input. Table~\ref{tab:metrics} shows the distribution of gain from the real data and the adversarial data. Table~\ref{tab:qualit} shows some qualitative examples. One demonstration where adversarial gain using the InferSent embedding space helps is with the third example in Table~\ref{tab:qualit}. Though the model's label changes, with a relatively small number of added words (2), the meaning of the sentence possibly changes indicating that ``William's best screen work'' may be cheesy. The shift in sentiment causes the adversarial gain to fall close to the gain of the real data and thus the model is less likely to be making a mistake if an oracle were to label the perturbed sample.

\section{Discussion}
Overall, we introduce the notion of adversarial gain as a measure of adversary effectiveness and model robustness against an adversary. This notion is applicable to both generative and discriminative models and bears particularly convenient properties for many tasks in natural language processing. While the notions of distance which we provide here are not perfect, they appear to provide adequate information to assess performance. In the future, learning a domain dependent feature representation space may help to improve the information provided by adversarial gain. Going forward, adversarial gain can provide a more standardized comparative measure of adversarial examples and attack quality. Furthermore, its roots in stability theory, use of manifold spaces, and other interesting properties as a unified view of adversarial examples may inspire the construction of future robust and gain-stable NLP models.

\putbib
\end{bibunit}

\onecolumn
\appendix
\label{sec:appendix}
\begin{bibunit}

\onecolumn
\section{Literature Review}

\begin{table}[H]
\footnotesize
\centering
\resizebox{\textwidth}{!}{
\vspace{-0.2\baselineskip}
\setlength\tabcolsep{3pt}
\begin{tabular}{p{-1.5ex}ll|lll|l|l}
\toprule
& & &
\multicolumn{3}{c|}{\textbf{Input Perturbation}} &
\multicolumn{1}{c|}{\textbf{Output}} & 
\multicolumn{1}{c}{\textbf{Task}}
\\[1ex]
&  Paper  & &
Input perturbation & Gradient & Use of Human \\& & & & Based & Perception & 
& 
 \\ 
\midrule
\multirow{7}{*}
& \multicolumn{2}{l|}{\citet{miyato2016adversarial}} & Perturbation to word embedding & Yes & No & Change in  &Text classification\\& & & &  &   & cost function &  \\
\midrule
& \multicolumn{2}{l|}{\citet{dasgupta2018evaluating}} & Change in word ordering & No & No & Change in class & Natural Language \\ & & & replace 'more' with 'less'
& & & &  Inference  \\
\midrule
& \multicolumn{2}{l|}{\citet{jia2017adversarial}} & Concatenate adversarial sentence  & No & Yes &  lower F1 score & Question \\ & & & replace 'more' with 'less'
& & & &  Answering  \\

\midrule

& \multicolumn{2}{l|}{\citet{samanta2017towards}} & Replace or remove words which   & No & No  &  Change in class & Sentiment \\ & & &contribute most to classification with  
& & & &  Analysis  \\
& & & synonyms, typos and genre specific words 
& & & &    \\
\midrule

& \multicolumn{2}{l|}{\citet{kuleshov2018adversarial}} & Replace word with synonym, & Yes & No &  Change in class & Classification \\ & & & decision learned using a constraint optimization. 
& & & &   \\

\midrule

& \multicolumn{2}{l|}{\citet{hosseini2017deceiving}} & Negating phrases, misspellings.
& No & No &  Lower toxicity score. & Classification \\ & & & decision learned using a constraint optimization. 
& & & &  on confidence  \\

\midrule

& \multicolumn{2}{l|}{\citet{cheng2018seq2sick}} & Non-overlapping exclusive words attack;
& Yes &  No &  Change in BLEU score & Translation \\ & & & targeted keyword attack where all the keywords 
& & & &  \\ & & &
must be present in the adversarial input; 
& & & & \\ & & &
word replacement. & & & & \\

\midrule

& \multicolumn{2}{l|}{\citet{papernot2016crafting}} & Replacing words with most impactful words
& Yes & No &  Change in class \& & Classification, \\ & & & in the classification w.r.t Jacobian quantity;
& & & distribution & Generation  \\
\midrule

& \multicolumn{2}{l|}{\citet{liang2017deep}} & Change characters, insert one hot word,
& Yes & Yes &  Change in cost function & Classification \\ & & &  parenthesis, forged fact
& & & &  \\
\midrule

& \multicolumn{2}{l|}{\citet{li2016understanding}} & Drops certain dimensions of word embeddings,
& Yes & No &  Change in class confidence & Classification \\ & & &   uses RL to find minimal set of words to remove
& & & &  \\
\midrule

& \multicolumn{2}{l|}{\citet{ebrahimi2017hotflip}} & Flips the characters/ words in a sentence w.r.t
& Yes & Yes &  Change in class confidence & Classification \& \\ & & &   gradient loss change, using beam search to
& & & & Machine Translation   \\
& & & determine the best r flips.
& & & & \\
\midrule

& \multicolumn{2}{l|}{\citet{gao2018black}} & 
Change in characters / words w.r.t
& No & No &  Change in class & Classification \\ & & &   
token importance & & & &   \\
\midrule

& \multicolumn{2}{l|}{\citet{zhao2017generating}} & 
Generate adversarial examples by using 
& Yes & Yes &  Change in class & Textual Entailment \& \\ & & &   
Adversarially regularized autoencoders. & & & & Machine translation   \\
\midrule

\bottomrule
\end{tabular}}
\caption{Definition used for previous work on adversarial examples}
\label{tab:adversarial_def}
\vspace{-1mm}
\end{table}

Recently, there has been many previous work done on adversarial examples in the text domain. Broadly speaking, the attacks can be categorized as gradient based and non-gradient based. For gradient based attacks the adversarial input is chosen based on change in cost functions and model gradients, which are also known as \emph{white-box} attacks for their ability to look into the model while constructing the adversarial input. Similarly, non-gradient based attacks rely on clever input manipulations such as misspellings, addition, removal or replacement of words keeping the same semantic meanings. These kind of attacks are also termed as \emph{black-box} attacks. We present a brief review over the existing works in Table \ref{tab:adversarial_def}. We provide an additional column on human perception, which denotes whether the paper has accounted for human perception of the attack in some way. That is whether the proposed attacks can be discerned from the original text by human annotators.

\subsection{Definitions}

Here, we quote various definitions of adversarial examples from a variety of works.

\medskip\noindent \textit{We expect such network to be robust to small perturbations of its input, because small perturbation cannot change the object category of an image. However, we find that applying an imperceptible non-random perturbation to a test image, it is possible to arbitrarily change the network’s prediction.}~\cite{szegedy2013intriguing}

\medskip\noindent \textit{That is, these machine learning models misclassify examples that are only slightly different from correctly classified examples drawn from the data distribution}~\cite{goodfellow2014explaining}

\medskip\noindent \textit{Adversarial examples are examples that are created by making small perturbations to the input designed to significantly increase the loss incurred by a machine learning model}~\cite{miyato2016adversarial}

\medskip\noindent \textit{Our goal is to design pairs of sentences such that the NLI relation within a pair (entailment, neutral or contradiction) can be changed without changing the words involved, simply by changing the word ordering within each sentence.}~\cite{dasgupta2018evaluating}

\medskip\noindent \textit{We define an adversary A to be a function that takes in an example $(p, q, a)$, optionally with a model $f$, and returns a new example $(p_0 , q_0 , a_0 )$. The adversarial accuracy with respect to $A$ is $Adv(f)= \frac{1}{|D_{test}|} \sum_{(p,q,a) \in D_{test}} v(A(p, q, a, f), f))$. While standard test error measures the fraction of the test distribution over which the model gets the correct answer, the adversarial accuracy measures the fraction over which the model is robustly correct, even in the face of adversarially-chosen alterations...Instead of relying on paraphrasing, we use perturbations that do alter semantics to build concatenative adversaries, which generate examples of the form $(p + s, q, a)$ for some sentence $s$. In other words, concatenative adversaries add a new sentence to the end of the paragraph, and leave the question and answer unchanged.}~\cite{jia2017adversarial}

\medskip\noindent \textit{An adversarial sample can be defined as one which appears to be drawn from a particular class by humans (or advanced cognitive systems) but fall into a different class in the feature space.}~\cite{samanta2017towards}

\medskip \noindent \textit{maliciously crafted inputs that are undetectable by humans but that fool the algorithm into producing undesirable behavior}~\cite{kuleshov2018adversarial}

\medskip \noindent \textit{Adversarial examples are inputs to a predictive machine learning model that are maliciously designed to cause poor performance}~\cite{ebrahimi2017hotflip}

\medskip \noindent \textit{One type of the vulnerabilities of machine learning algorithms is that an adversary can change the algorithm output by subtly perturbing the input, often unnoticeable b y humans.}~\cite{hosseini2017deceiving}

\medskip \noindent \textit{Adversarial attack on deep neural networks (DNNs) aims to slightly modify the inputs of DNNs and mislead them to make wrong predictions}~\cite{cheng2018seq2sick}

\medskip \noindent \textit{For a given sample x and a trained DNN classifier model F, the attacker aims to craft an adversarial sample x* = x + x by adding a perturbation x to x, such that F(x*)  F(x)...In order to maintain the utility of a text sample, we perturb the sample not only by directly modifying its words, but also inserting new items (words or sentences) or removing some original ones from it.}~\cite{liang2017deep}

\subsection{Adversarial Gain Perspectives of Prior Work}

Here we examine various works and how they can fit into the adversarial gain perspective. We already demonstrate how~\cite{cheng2018seq2sick} and~\cite{ebrahimi2017hotflip} can be measured in terms of adversarial gain. Rather than non-overlapping text in~\cite{cheng2018seq2sick}, we can examine the semantic change of the output. Similarly, we can examine how well the noise preserves the meaning of the input sentence in both cases. If the semantic shift is too far, this discounts the shift in output it causes.

Generally, most text-based adversarial attacks constrain their inputs by in some way changing words while retaining meaning. This includes negation~\cite{hosseini2017deceiving}, misspelling~\cite{hosseini2017deceiving,samanta2017towards,liang2017deep}, changing word order~\cite{dasgupta2018evaluating}, replacing with with synonyms~\cite{cheng2018seq2sick}, or simply perturbing the word embeddings~\cite{miyato2016adversarial}. In many cases, such constraints will preserve the meaning of the original text, but often too strict constraints can result in lower success. For example, in~\cite{ebrahimi2017hotflip} the word-based replacement with a strict synonym constraint resulted in a low success rate in adversarial examples. In other cases, preservation of word meaning is not guaranteed. In fact, prior work has used samples from the generated attacks posed as surveys to determine whether meaning is preserved~\cite{jia2017adversarial}, but this has not typically been done in a systematic way and \citet{jia2017adversarial} found that in some cases meaning was not preserved. In another example, negation of phrases does not preserve meaning and thus a model could be totally correct in changing its output. In all attacks, it is possible to evaluate preservation of meaning by using a well-defined embedding space (such as~\cite{conneau2017supervised} as a start) and the cosine distance. The use of such a distance as we do as part of adversarial gain, allows attacks to change meaning and account for this when evaluating the change of the model output. 

In evaluating the results of an adversarial attack, there are many measures used. For classification tasks, a change in class label is typically used as a success criterion~\cite{zhao2017generating,gao2018black,kuleshov2018adversarial,samanta2017towards,dasgupta2018evaluating}. In other cases, notions such as changes in some scoring or cost function are used~\cite{cheng2018seq2sick,jia2017adversarial,hosseini2017deceiving,miyato2016adversarial}. However, due to the change in inputs if the cost relates to the original sentence and the meaning is \emph{not preserved}, the cost may be evaluating the wrong criterion without access to an oracle. Thus adversarial gain accounts for this by discounting output performance changes by the distance from the input. In a well defined feature space where inputs and outputs are correlated this ensures that as an adversarial input moves away from its original meaning, this is accounted for in the evaluation criteria to some extent. 

\section{Extended Perspectives on Adversarial Gain}

Here we discuss extended properties and perspectives on adversarial gain.

\subsection{Possible Feature Spaces and Distance Metrics to Measure Gain}

There are a number of different priors that can be used to measure gain in different ways for different tasks. While in the main text we examine sentiment classification tasks and text summarization, others may be relevant in domains such as dialogue systems. For example, one can use sentiment classification probability and the likelihood divergence (or a step function intersection) to measure difference in output of a dialogue system, text summary system, or other generative model. Similarly, various sentence embeddings can be used (Infersent, Doc2Vec, etc.). Word-wise word vector distance can also be used. Each of these notions of adversarial gain essentially provide a different prior on the stability of the systems in different ways. For example, it is likely that unless the sentiment of an input to a dialogue system doesn't change dramatically, neither should the output.

\section{Experimental Setup}

In our selection of text-based attacks, we examined which attacks provided easily available open-source code. Many code to replicate experiments was either unavailable or we were unable to find. We settled on two text-based attacks. We used the Seq2Sick attack on text summarization by~\citet{cheng2018seq2sick} and the word-level sentiment classification attack by~\citet{ebrahimi2017hotflip}. Scripts and full instructions that we used to run the code from these papers is provided at: \href{anonymized}{anonymized}. More samples with gain and distances provided can be found in the codebase provided.
\subsection{Text Summarization}

We use the pre-trained model and code for a text summarization model based on the Open Neural Machine Translation toolkit (OpenNMT)~\cite{2017opennmt} as provided by~\citet{cheng2018seq2sick} at \href{https://github.com/cmhcbb/Seq2Sick}{https://github.com/cmhcbb/Seq2Sick}. We use the GigaWord corpus the authors reference from~\cite{rush2015neural} based on prior versions of the datset~\cite{graff2003english,napoles2012annotated}.

When we measure cosine distance, we use the inverse of cosine similarity to follow the intuition the a distance metric should keep similar words closer together. Assuming that cosine similarity is bounded $[0,1]$, cosine distance is $1-|similarity(x,y)|$.

\subsection{Sentiment Classification}

For sentiment classification we use the binary version of the SST dataset~\cite{socher2013recursive} called SST2. This removes all neutral labels. This is the same dataset as used by~\cite{ebrahimi2017hotflip}. We use their provided code for the word-level adversarial attack and SST2 pre-processing scripts found at: \href{https://github.com/AnyiRao/WordAdver}{https://github.com/AnyiRao/WordAdver} and \href{https://github.com/AnyiRao/SentDataPre}{https://github.com/AnyiRao/SentDataPre}. We use the pre-trained convolutional neural network classification model provided by the authors and the attack as provided in our accompanying instructions. The only change we make is that we remove the cosine similarity requirement on replacement words. We do this because otherwise the attack only generates attacks for 95 samples. Removing this requires generates attacks for all samples (though many are not successful). We note that this allows words to be added by replacing padding characters, while this differs slightly from the attack mentioned by~\cite{ebrahimi2017hotflip}, the authors there do discuss that this attack has a low success rate particularly due to their restrictions. Because adversarial gain as a definition does not require constraints, this allows us to consider the larger set of attacks.

\putbib
\end{bibunit}

\end{document}